\documentclass[10pt, conference, compsocconf]{IEEEtran}
\usepackage {mathtools}
\usepackage{graphicx}
\usepackage{subfig}
\usepackage{color}
\usepackage{textcomp}
\usepackage{booktabs}
\usepackage{amsmath,amsfonts,amssymb} 
\usepackage{cite}
\graphicspath{{}} 
\IEEEoverridecommandlockouts                              

\title{\LARGE \bf
	Automatic Calibration of Dual-LiDARs Using Two Poles Stickered with Retro-Reflective Tape
}

\author{Bohuan Xue$^{1}$, Jianhao Jiao$^{1}$, Yilong Zhu$^{1}$, Linwei Zhen$^{1}$, Dong Han$^{2}$,  Ming Liu$^{1}$, Rui Fan$^{1}$

	\thanks{$^{1}$ Bohuan Xue, Jianhao Jiao, Yilong Zhu, Linwei Zhen, Ming Liu and Rui Fan are with the Robotics and Multi-Perception Lab, Robotics Institute, Hong Kong University of Science and Technology, Hong Kong SAR, China. \{bxueaa, jjiao, yzhubr, lzhengad, eelium, eeruifan\}@ust.hk}%
	\thanks{$^{2}$ Dong Han is with the Shenzhen Institutes of Advanced Technology, Chinese Academy of Sciences, Shenzhen, China. dong.han@siat.ac.cn}%
}

\begin{document}

	\maketitle

	\begin{abstract}
		
    Multi-LiDAR systems have been prevalently applied in modern autonomous vehicles to render a broad view of the environments. 
	The rapid development of 5G wireless technologies has brought a breakthrough for current cellular vehicle-to-everything (C-V2X) applications. Therefore, a novel localization and perception system in which multiple LiDARs are mounted around cities for autonomous vehicles has been proposed.
	However, the existing calibration methods require specific hard-to-move markers, ego-motion, or good initial values given by users.
	In this paper, we present a novel approach that enables automatic multi-LiDAR calibration using two poles stickered with retro-reflective tape. This method does not depend on prior environmental information, initial values of the extrinsic parameters, or movable platforms like a car.
	We analyze the LiDAR-pole model, verify the feasibility of the algorithm through simulation data, and present a simple method to measure the calibration errors w.r.t the ground truth. 
	Experimental results demonstrate that our approach gains better flexibility and higher accuracy when compared with the state-of-the-art approach.
	\end{abstract}


	\section{INTRODUCTION} \label{sec0}
	
    Unmanned driving technology is becoming more and more popular \cite{brink2017bits}. Nowadays, 5G technology is accelerating the development of cellular vehicle-to-everything (C-V2X) technology \cite{fan2018real}, in which unmanned vehicles need to perceive various objects to navigate and avoid obstacles \cite{fan2017real}. In such C-V2X systems, in addition to cameras, LiDARs are used because illumination can affect the image quality of \mbox{cameras \cite{zhu2019real}}, and the position estimation of feature points is related to the accuracy of cameras' intrinsic and extrinsic parameters \cite{pereira2016self}. 	
	However LiDARs have several limitations. Firstly, LiDARs have blind areas \cite{fan2019key}. For example, if vehicles are surrounded by tall trucks, they will lose most of the observation information. 
	Locating an unmanned system requires the point cloud information around the vehicle, and the point cloud based location method has several shortcomings. For various reasons, the prior 3D surfel maps may change — for example, road construction or vegetation pruning at the roadside. Secondly, LiDARs are very expensive, and multi-LiDAR solutions are costly \cite{zheng2019low}.

\begin{figure}[t]
		\centering
		\includegraphics[width=0.30\textheight]{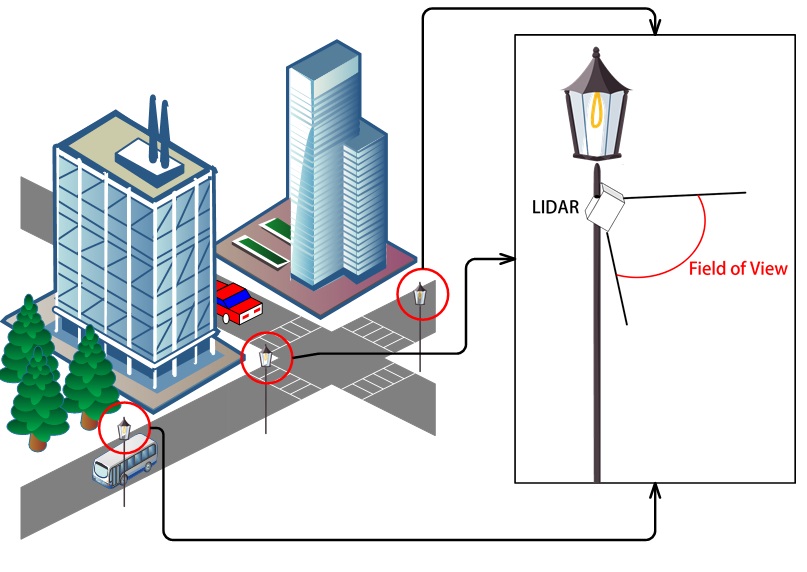}
		\caption{LiDARs mounted on lampposts. Because the LiDARs are mounted high, they can provide additional information on the ground. Moreover, multiple LiDARs can provide fewer blind spots.}
		\label{figure-building}
		\vspace{-1.5em}
		\end{figure}

	Mounting LiDARs on lampposts, as shown in Fig. \ref{figure-building}, can solve previously problems. Lamppost LiDARs can provide a real-time surfel map, so moving vehicles can directly receive 3D information provided by the lamppost through the 5G network.


	
	But there is a problem that the extrinsic parameters of LIDARs mounted on lampposts or other urban facilities are unknown, and we need to know their position in the world to obtain the complete 3D city real-time surfel map. To solve these problems, the calibration of a multi-LiDAR configuration is necessary. 
	 
	Over the past years, many methods for calibrating LiDARs have been proposed, but several drawbacks are presented. Several of such methods rely exclusively on an additional sensor, need a good initialization provided by users, or need complex objects such as walls, which may not exist on docks and in other scenarios. Motion-based approaches require LiDARs to be installed on mobile platforms such as cars. So these methods are not feasible in the case shown \mbox{in Fig. \ref{figure-building}}. For example, tracking-based methods need moving objects to be tracked, but the accuracy of tracking affects the calibration result, and such methods require significant human intervention.
	

    In this paper, we propose an automatic calibration approach for the proposed multi-LiDAR system. First, we physically lay them down on the horizontal surface on which you walk, i.e, the ground. After that, we extract the poles from LiDAR data using intensity information. Then we use the identified poles to construct constraints so that the calibration problem is transformed into an optimization problem. Finally, we provide a way to choose the correct results, because there are several locally optimal solutions. 	
	We conduct a variety of experiments to show the reliability and accuracy of our proposed approach. The contributions of this paper are summarized as follows:
	\begin{itemize}
		\item We propose a simple method which calculates the position of points on arbitrary poles and the points are generated by a LiDAR. Then the errors can be theoretically analyzed.
		\item We give a method of calibrating  LiDARs using two non-parallel poles stickered with retro-reflective-tape. The themod can be used to obtain a good result, and can be used in scenarios where LiDARs can not be moved as described in Fig. \ref{figure-building}.
		\item We provide extensive evaluation experiments on simulated and real-world datasets.
	\end{itemize}
	
	The remainder of the paper is organized as follows. \mbox{Section \ref{sec1}} gives a review of related works. The methodology of our approach is described in Section \ref{seg3}. Implementation and tests are shown in Section \ref{seg4}. Finally, Section \ref{seg5} concludes this work.
	
	\section{RELATED WORK}  \label{sec1}
	There are two main types of LiDAR calibration approaches, appearance-based and motion-based. 
	The former type of approaches fall into turning problems into registration problems, and the approaches usually need fixed markers or prior environment. The latter type of approaches utilize the constraints of sensors' motion to recover the extrinsic parameters. Then the approaches are formulated as the well-known hane-eye calibration problem. 
	The accuracy of the results of such approaches is related to the accuracy of estimated motions, which is easily affected by accumulated drifts.

	Underwood et al. \cite{underwood2007calibration} propose a calibration method that needs one vertical pole with retro-reflective tape and a sensor platform limited to a planar surface. The platform must be moved so that the sensos can observe the environment from different headings. 
	Gao et al. \cite{gao2010line} use retro-reflective targets placed in scenes to calibrate a multi-LiDAR system, and this approach needs the position of the vehicle platform and the initial calibration estimate. 
	All these approaches need a platform that can be moved, so they are hard to apply in a LiDAR fixed system, like that in Fig. \ref{figure-building}. 
	
	Shang et al. \cite{shang2015efficient} present a calibration method for 3D LiDARs, which only needs an orthogonal normal vector pair, and the normal vector needs to be generated from a planar ground plane and a vertical wall.  \mbox{Jiao et al.\cite{jiao2019novel}} use three linearly independent planar surfaces to find correspondences to enable automatic LiDAR calibration, but the requiement of three planar surfaces is very demanding.
	\mbox{Muhammad et al. \cite{muhammad2010calibration}} propose a method for multi-beam LiDARs. This technique is based on an optimization process performed to estimate the LiDAR calibration parameters from a coarse initial calibration. The drawback of all these calibration approaches is that they need a specific environment, a mobile platform or a reliable initial value, which are hard to get in some situations.
	Our approach only depends on two non-parallel poles stickered with retro-reflective tape, and these are easy to place in any situation. We do not need the LiDAR platform to be movable, nor do we need an initial calibration estimate, which makes our approach more general and practical. 
	
	Many LiDAR-camera calibration algorithms have emerged. Levinson et al. \cite{levinson2013automatic} introduce techniques that enable camera-laser calibration online, automatically and in arbitrary environments, using a probabilistic monitoring algorithm. \mbox{Martin et al. \cite{martin2014calibration}} present a pipeline for mutual pose and orientation estimation of the LiDAR-camera system using a coarse-to-fine approach. Pandey et al.\cite{pandey2012automatic} report on a mutual information (MI) algorithm. MI as the registration criterion can work in a situation without the need for any specific calibration targets.
	
	
	
	Motion-based approaches all require that the LiDARs can be moved. Heng et al. \cite{heng2013camodocal} publish a tool called CamOdoCal, a versatile algorithm which does not need any prior knowledge about the rig setup. 
	Jiao et \mbox{al. \cite{jiao2019automatic}} align the estimated motions of each sensor as an initialization then refine them with an appearance-based method. In our cases shown, in Fig. \ref{figure-building}, motion-based approache are all impossible, because the LiDARs can't be moved. 
	Quenzel et al. \cite{quenzel2016robust} present a method, using pose graph optimization to calibrate the extrinsic parameters of LiDARs. However this method needs one object to be clustered exactly into one segment.
	
	Many of the above techniques rely on one or more assumptions and are not applicable in our case. 
	Our approach only needs two poles with retro-reflective tape that helps us to recognize the poles. It doesn't involve ego-motion nor an environment prior assumption. We only need to place the LiDARs non-parallel in the LiDARs' overlapping area.


	\section{METHODOLOGY} \label{seg3}
	We first place two poles in a position in which they are not parallel to each other. In this section, we provide details of the process.
	
	\subsection{Pole Extraction and Representation} 
	Because the poles have been stickered with retro-reflective tape, it is easy to identify them from the point cloud using a simple threshold operation with the parameter of intensity. When the distance between the LiDAR and pole is \mbox{about 5 meters}, the parameters can be chosen from those that are presented in \mbox{Tab. \ref{parameter_selection}}. 
	
	\begin{table}[t]
		\caption{Parameter Selection of Different LiDARs}
		\label{parameter_selection}
		\begin{center}
			\begin{tabular}{cc}
				\hline \toprule[0.03cm]
				LiDAR Manufacturers& Threshold of Intensity \\
				\hline \toprule[0.03cm]
				Velodyne & 230\\
				Hesai & 200\\
				Leishen & 200\\
				RoboSense & 200\\
				\hline \toprule[0.03cm]
			\end{tabular}
		\end{center}
			\vspace{-2em}  
	\end{table}
	
	Even though we filter out many irrelevant points by the threshold, there are some outliers. These points can be filtered by using an arbitrary clustering algorithm because their numbers are small in the majority of cases. 
	
	The pole point cloud $\mathcal{P}$ can be represented by a linear equation, and the linear equation can be denoted as a vector equation: $\mathbf{r}=\mathbf{p}+\lambda\mathbf{n}$ where $\lambda$ is a scalar. Then we denoted it as $(\mathbf{p}, \mathbf{n})$, where $\mathbf{p}$ means the line through the \mbox{point $\mathbf{p}=[p_x, p_y, p_z]^\top$} and $\mathbf{n}=[n_x, n_y, n_z]^\top$ denotes the direction of the line. Another way of expressing the pole point cloud is to use $\mathcal{P}=\{\mathbf{p_1,p_2,...}\}$, where $\mathbf{p_i}$ means the $i^{th}$ point in the point cloud. For convenience, $L_{i}$ is used to represent the transformation from the world to the $i^{th}$ LiDAR frame, and $\mathcal{P}_{ij}$ to represent the pole $j$ captured by the $i^{th}$ LiDAR.

	\subsection{Initialization for Calibration}
\begin{figure}[t]
		\centering
		\includegraphics[width=0.18\textheight]{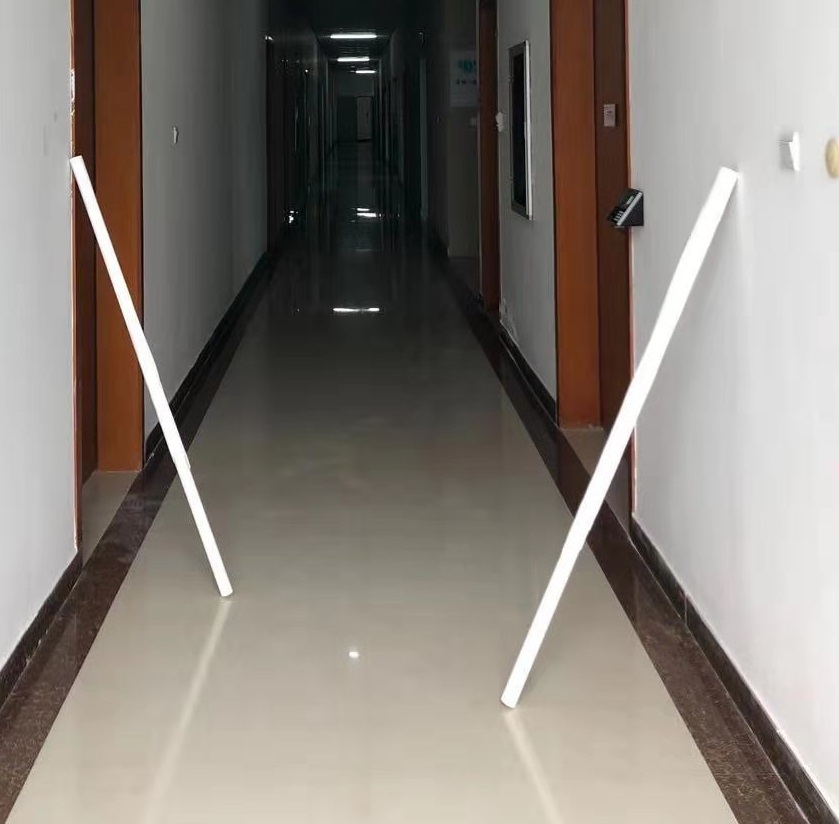}  
		\caption{The poles stickered with retro-reflective tape.}
		\label{poles_under_ground}
	\end{figure}

\begin{figure}[]
		\centering
		\includegraphics[width=0.31\textheight]{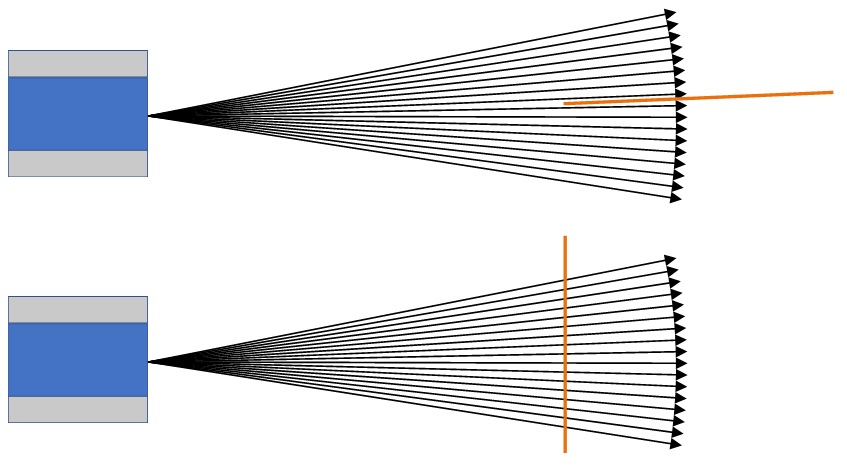}  
		\caption{The LiDAR beam and pole. The orange line indicates a pole. The black line indicates the beam of LiDAR. A horizontal pole can cause the beams to fail to reach the pole itself. Placing the pole vertically gets more useful data.}
		\label{lidar_beam}
				\vspace{-1em}
	\end{figure}

	As shown in Fig. \ref{poles_under_ground}, we can place the poles arbitrarily. Considering the type of LiDAR, such as VLP16\footnote{Velodyne 16-channel mechanical LiDAR is one of the most common LiDARs .} or Pandar64\footnote{Pandar64 is a 64-channel mechanical LiDAR from Hesai Technology.}, we should not arrange the pole horizontally because the LiDAR's beams may not scan the poles, as can \mbox{be seen in Fig. \ref{lidar_beam}}.
	
	LiDAR calibration is accomplished by aligning the poles from different LiDARs. Essentially, we want to get a rotation matrix of $\mathbf{R}$ and a translation vector of $\mathbf{t}$, which describes the pose relationship between the two LiDARs. We represent the two LiDARs in two different ways, $\mathcal{P}_1=\{\mathbf{p_1,p_2,...}\}$ and $\mathbf{r}_1=\mathbf{p}+\lambda\mathbf{n}$ respectively. Then these data are acquired by solving a least-squares problem:
	\begin{equation}
	\label{equation_arg_min}
	\mathbf{R}^*, \mathbf{t}^* =\mathop{\arg\min_{\mathbf{R}, \mathbf{t}}} \ \ \sum\limits_{n = 1}^N {\|{\mathbf{p_n}} - \mathbf{r}_1 \|},
	\end{equation}
	where $\left\|\cdot\right\|$ indicates the $\ell_{2}$-norm of a vector. For the above case, there is only one line $\mathbf{r}_1$ and one point cloud $\mathcal{P}_1$, so we can get infinite solutions. If there are two lines and two point clouds, and they are not parallel, the number of solutions will be greatly reduced. So we introduce the variables $\pi_{n,k}$. If the point cloud $n$ corresponds to line $k$, then the $\pi_{n,k}$ will \mbox{equal 1}, else it will equal to 0. We subsequently slightly improve (\ref{equation_arg_min}) to get
	
	\begin{equation}
	\label{equation_arg_min2}
	\mathbf{R}^*, \mathbf{t}^* =\mathop{\arg\min_{\mathbf{R}, \mathbf{t}}} \ \ \sum\limits_{n = 1}^N \pi_{n,k}{\|{\mathbf{p_n}} - \mathbf{r}_1 \|}.
	\end{equation}

	\subsection{Determination of Correspondence Relation}
	
\begin{figure}[t]
		\subfloat[]{%
			\includegraphics[width=0.09\textwidth]{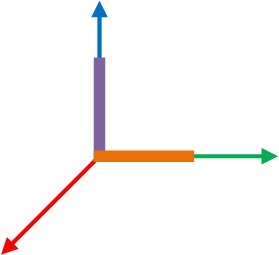}}
		\qquad
		\subfloat[]{%
			\includegraphics[width=0.09\textwidth]{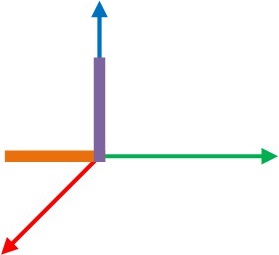}}
		\qquad
		\subfloat[]{%
			\includegraphics[width=0.09\textwidth]{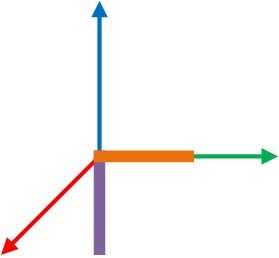}}
		\qquad
		\subfloat[]{%
			\includegraphics[width=0.09\textwidth]{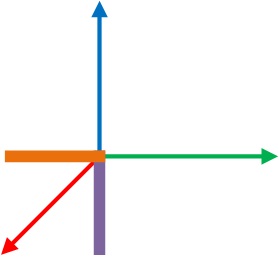}}
		
		\centering
		\caption{Four different correspondence relationships. The red, blue, and green lines represent three directions of a coordinate axis. We assume that the green and the blue direction are the two poles' directions. The ground truth is (a). By rotating and transforming the ground truth, three other convergent results can be obtained.}
		\label{four_axis}
				\vspace{-1.5em}
	\end{figure}

	If we have an excellent prior, it is easy to determine the correspondence relationship $\pi_{n,k}$, but in many cases, it is not easy to get the prior. Although we can get a reliable initial pose by adjusting the points using editing tools like \mbox{Cloudcompare \cite{girardeau2011cloudcompare}}, it is complicated to do this. Each pairing relationship has four different solutions, and there are two different correspondence relationships, so there are eight different situations. Fig. \ref{four_axis} illustrates four situations which we can match. 
	
	We can enumerate all the corresponding relationships and then find a reasonable result.
	Although it is easy for human beings to judge what is a reasonable result, for a computer, this is not easy because it does not have enough semantic information to find the correspondence. There is a solution in which we can use the Iterative Closest Point (ICP) \mbox{algorithm \cite{Pomerleau2013Comparing}} to register the two point clouds, and choose the result that is the closest to the identity. The result of rotation is $\mathbf{R}_{result}$, and the translation the result is $\mathbf{t}_{result}$, and the error in rotation can be represented as $e_r=\left\|\log(\mathbf{R}_{result})^{\vee}\right\| $, where $\log(\cdot)^\vee$ is defined to transform $\textrm{SO(3)}$ to $\mathbb{R}^3$. Similarly, the error in translation is $e_t=\left\|\mathbf{T}_{result}\right\|$. In most situations, the correct result's rotation and translation error is always minimal. The relevant results can be seen in \mbox{Sect. \ref{seg4}}.
	
	
	\subsection{Accuracy Evaluation}
	We next solve the problem of why a pole can be represented by a line and how to measure the error. First, we can assume that the pole is placed vertically at the origin of the coordinate axis, and the pole's central axis is the Z-axis. The LiDAR is put in the X-axis with arbitrary orientation.
	The parameters are only the LiDAR's position $\mathbf{t}=[x_p, 0, 0]^\top$ and direction, which can be represented as a vector. The vector is vertical to the central plane of the LiDAR, so the vector can be represented by $\mathbf{v}=[x_n, y_n, z_n]^\top$, and the radius of the pole is $r$. Then we can get the equation of the cylinder:
	\begin{equation}
	\label{cylinder_equation}
	\begin{aligned}
	x &= r cos(\theta),\\
	y &= r sin(\theta),\\
	z &= -\big[x_n r cos(\theta) - x_l x_n + y_n r sin(\theta)\big]/z_n.
	\end{aligned}
	\end{equation}
	
\begin{figure}[t]
		\subfloat[]{%
			\label{figure_matlab_a}
			\includegraphics[width=0.28\textwidth]{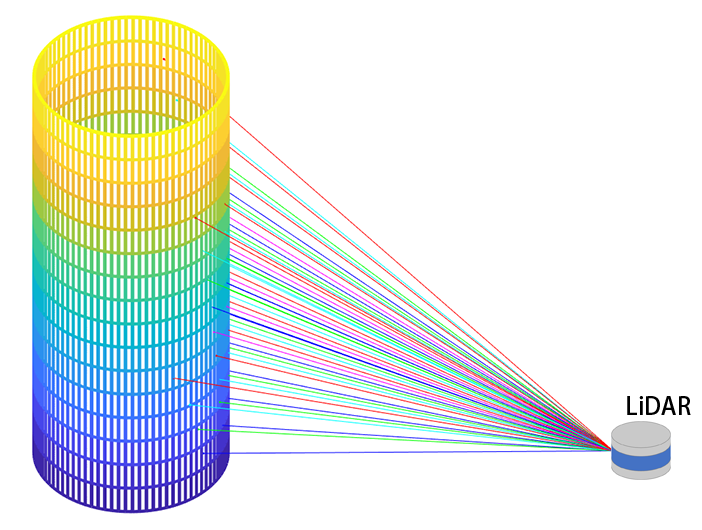}}
		\qquad
		\subfloat[]{%
			\label{figure_matlab_b}
			\includegraphics[width=0.28\textwidth]{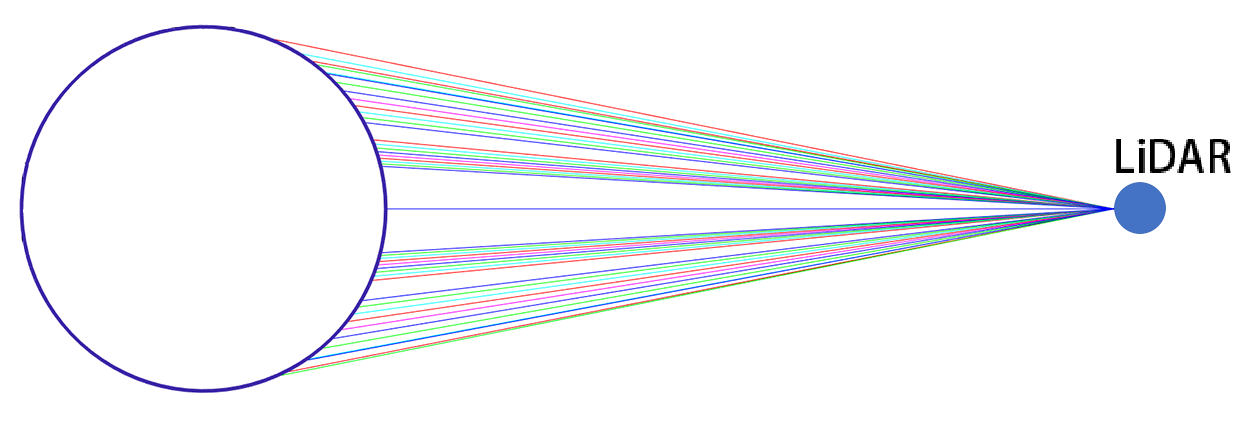}}
		\centering
		\caption{The points in the cylinder. The coloured lines are the LiDAR's beams, which touch the cylinder. The LiDAR's pose can be represented as a quaternion $[0.884,0.306,0.177,0.306]^\top$. \mbox{Fig. (a)} the normal perspective, from which we can see the points in the cylinder. \mbox{Fig. (b)} is the top view, all points fall precisely on the edge of the cylinder.}
		\label{two_matlab}
				\vspace{-1.5em}
	\end{figure}

	In addition, considering the pole is a cylinder, it can obtain a range of the value $\theta = \arccos(r/x_p)$. Then, if we get a point on the pole and its coordinate parameter is $\theta$ in (\ref{cylinder_equation}), we can get  
	\begin{equation}
	\label{cylinder_tangent_equation}
	\begin{aligned}
	\left( {f(\theta) - \mathbf{t}} \right)\cdot \left( {f(\theta  + d\theta ) - \mathbf{t}} \right) =\\ \cos(c)\left\| {f(\theta ) - \mathbf{t}} \right\|\left\| {f(\theta  + d\theta ) - \mathbf{t}} \right\|,
	\end{aligned}
	\end{equation}
	where the function $f$ corresponds with (\ref{cylinder_equation}), $\mathbf{t}$ is the LiDAR's position mentioned above, and $c$ is the angular resolution parameter of each beam. When the information is sufficient, it is easy to get the value of the variables $d\theta$ using an algorithm to solve nonlinear equation systems, like the preconditioned conjugate \mbox{gradients \cite{berry1994templates}}, the trust-region-dogleg \mbox{algorithm \cite{powell1970fortran}} or the Levenberg-Marquardt \mbox{method \cite{levenberg1944method}, \cite{marquardt1963algorithm}}. Therefore if we have a current LiDAR position and its beam parameters, we can use (\ref{cylinder_tangent_equation}) to infer the coordinates of all points on the pole. The result can be seen in \mbox{Fig. \ref{two_matlab}}.
	
	When we get all the points on a pole, these points can be fitted by a straight line, and the line is similar to the central axis of the pole. The fitted line can be represented as a vector 
	$\left[n_{cx},n_{cy},n_{cz}\right]^\top$ 
	and a point $\left[x_c,y_c,z_c\right]^\top$ on the line. We can use
	\begin{equation}
	\label{measure_fit}
	{{{n_{cz}}} \over {{z_{\max }} - {z_{\min }}}}\int_{\left( {{z_{\min }} - {z_c}} \right)/{n_{cz}}}^{\left( {{z_{\max }} - {z_c}} \right)/{n_{cz}}} {{{\left( {{x_c} + {n_{cx}}z} \right)}^2}{{\left( {{y_c} + {n_{cy}}z} \right)}^2}dz},
	\end{equation}
	to measure the error between the fitted line and the central axis of the pole, where $z_{max}$ and $z_{min}$ are the points with maximum and minimum Z-coordinates on the pole.
	
	The smaller the result in (\ref{measure_fit}), the better the fitted result. Because the VLP-16 is a common LiDAR, we choose its parameters, so $c=0.2^\circ$. Quaternions can be used to indicate the orientation of LiDAR. If the quaternion is notated as $\mathbf{q}=[q_w,\ q_x,\ q_y,\ q_z]^\top$, then we provide the following results for reference: $\mathbf{q}_1=\left[0.957,\ -0.120,\ 0.263,\ -0.013\right]^{\top}$ and $\mathbf{q}_2=\left[1,\  0,\ 0,\ 0\right]^{\top}$.

	\begin{table}[t]
		\caption{Error from Different Parameters}
		\label{error1_table2}
		\begin{center}
			\begin{tabular}{c c c c}
				\hline \toprule[0.03cm]
				$x_p$ & $r$ & Error($\mathbf{q}=\mathbf{q_1}$) & Error($\mathbf{q}=\mathbf{q_2}$) \\
				\hline \toprule[0.03cm]
				10 & 0.3 & 0.061196 & 0.060038 \\
			
				10 & 0.2 & 0.027408  & 0.028638 \\
			
				10 & 0.1 & 0.007277  & 0.008005 \\
				
				6 & 0.3 & 0.059603  &  0.058643\\
			
				6 & 0.2 &  0.026762 &  0.026904\\
			
				6 & 0.1 &  0.006515  & 0.007360 \\
		
				4 & 0.3 & 0.061122  &  0.060220 \\
			
				4 & 0.2 & 0.026510  & 0.026064 \\
			
				4 & 0.1 & 0.006632  &  0.006237 \\
				\hline \toprule[0.03cm]
			\end{tabular}
		\end{center}
			\vspace{-2em}
	\end{table}
	
	It can be seen in Tab. \ref{error1_table2} that the smaller the value of $r$, the higher the accuracy. Considering the convenience, we will use a pole with $r=0.02$ in our later experiments.

	
	\section{EXPERIMENT} \label{seg4}
	
	In this section, we divide the evaluation into two separate steps. The initial calibration experiments are presented with simulated data and real sensor sets. Then we test the calibration approach on the real sensor data.
	
	\subsection{Experiments on Simulated Data}
	First, we assume that the positions of the poles and LIDARs are known. Then the calibration results are calculated with theoretical data using the approach mentioned above, and are compared with the ground truth. 

	We assume that the $i^th$ LiDAR $L_i$ and a pole's parameters $\mathbf{p}$, $\mathbf{n}$ and radius are known in the world frame. The task is to get the points in the LiDAR's coordinate system.
	
	First, we move the pole to the origin of the coordinate axis of the world frame. Then we get the new LiDAR $L^*$:
	\begin{equation}
L^*=\begin{bmatrix}
\mathbf{A} & 0\\ 
0 & 1
\end{bmatrix}
\begin{bmatrix}
\textrm{I} & \mathbf{-\!p}\\ 
0 & 1
\end{bmatrix}
L_i,
	\end{equation}
	where

	\begin{equation}
	\mathbf{a} = \frac{{{{\left[ {\begin{array}{*{5}{c}}
						{{-\!n_x}},\ {{-\!n_y}},\ {{1\!-\!n_z}}
						\end{array}} \right]}^T}}}{{\left\| {\left[ {\begin{array}{*{5}{c}}
					{{-\!n_x}},\ {{-\!n_y}},\ {{1\!-\!n_z}}
					\end{array}} \right]} \right\|}},
	\end{equation}
	\begin{equation}
				\mathbf{A}=2\mathbf{a}\mathbf{a}^\top-\textrm{I}.				
	\end{equation}
	We use $\mathbf{v}^*$ as the translation vector of $L^*$.
	Then, we put the LiDAR in the X-axis of the pole coordinate system get LiDAR $L$:
	\begin{equation}
		L=
		\begin{bmatrix}
		\textrm{I} & -\!\mathbf{v}^*_z\\ 
		0 & 1
		\end{bmatrix} L^*,
	\end{equation}
	where $\mathbf{v}^*_z=[0\ 0\ v^*_z]^\top$. Denoting $\mathbf{v}$ as translation vector of $L$, we can get the final LiDAR $\hat{L}_i$:
	
	\begin{equation}
		\hat{L}_i=\begin{bmatrix}
		\mathbf{R} & 0\\ 
		0 & 1
		\end{bmatrix} LL_i^{-1},
	\end{equation}
	where	
	\begin{equation}
	\mathbf{R}=2\mathbf{b}\mathbf{b}^\top-\textrm{I},				
\end{equation}

\begin{equation}
\mathbf{b}=\frac{%
	\left[
	\begin{matrix}%
		v_x\!+\!\left\|\mathbf{v}\right\|,\ v_y,\ v_z%
	\end{matrix}
	\right]}
{
		\left\| 
	\left[\begin{matrix}%
v_x\!+\!\left\|\mathbf{v}\right\|,\ v_y,\ v_z%
\end{matrix}
\right]
		\right\|
	}.
	\end{equation}

	
	
	Now we can get one LiDAR's points in the pole coordinate system using the method mentioned in the above section, and then we put these points into the LIDAR coordinate system. If we have two LiDARs, $L_1$ and $L_2$, and $L_1$ as a reference, the ground truth is $L_1^{-1}L_2$.
	
	We randomly generate parameters of the two poles and the postion of the LIDARs. Each pole's z-component of the directinal vector is larger than 0.9. If we think of the pole as a line, the distance bewteen the two poles is the distance between two points $\mathbf{p}_a$ and $\mathbf{p}_b$, which are in their line position of $z=0$. And the distance between the points is between 1.5 and 4. The positions $\mathbf{p}_a$ and $\mathbf{p}_b$ are in the y-axis and $2\leq\left\|\mathbf{p}_a\right\|, \left\|\mathbf{p}_b\right\|\leq3$. For the parameters of LiDAR position, the LiDARs are fixed, their positions are $[1,-\!1,0]^\top$ and $[1,1,0]^\top$, and their oritentaions are reprenseted as quaternions $[0.988,0.094,0.079,0.094]^\top$ and \mbox{$[0.989,-\!0.079,-\!0.094,-\!0.079]^\top$}. 
	
	For tests, we perform ten trials on the noisy data and compute the mean as well as the standard deviation of the rotation and translation error. Each LiDAR's points are subjected to zero-mean Gaussian noise with a standard deviation of 0.006 \cite{kidd2017performance}.
	
	To measure the difference in our results and ground truth, we can use the formula:
	\begin{equation}
	\label{measure_error}
	{e_{rt}}=\frac{1}{m-n}\int_n^m {{{\left\|{ {\mathbf{R}_{res}}{\mathbf{p}_x}+\mathbf{t}_{res}-{\mathbf{R}_{gt}}{\mathbf{p}_x}-{\mathbf{t}_{gt}}}\right\|}}dx}, 
	\end{equation}
	where $P_x=[0,\ x,\ 0]^T$, $n$ and $m$ represent the nearest and farthest distance of LiDAR in use. So we can assume that $n=1$ and $m=60$. Therefore, (\ref{measure_error}) describes the average error in using the LiDAR. The estimated extrinsic parameters of this algorithm are quite close to the ground truth; the $e_{rt}=0.168$. This proves that the proposed method can successfully calibrate the extrinsic parameters.
	
	\subsection{Experiments on Real Data}
	
	We calibrate a sensor system that consists of two LiDARs in a corridor. Since the precise extrinsic parameters of the system are unknown, and it's impossible to get its accurate value, we use Rényi's \mbox{Quadratic Entropy (RQE)\cite{sheehan2012self}} to evaluate our results.
	
	\begin{figure}[t]
		\subfloat[]{%
			\label{fig.result_a}
			\includegraphics[width=0.225\textwidth]{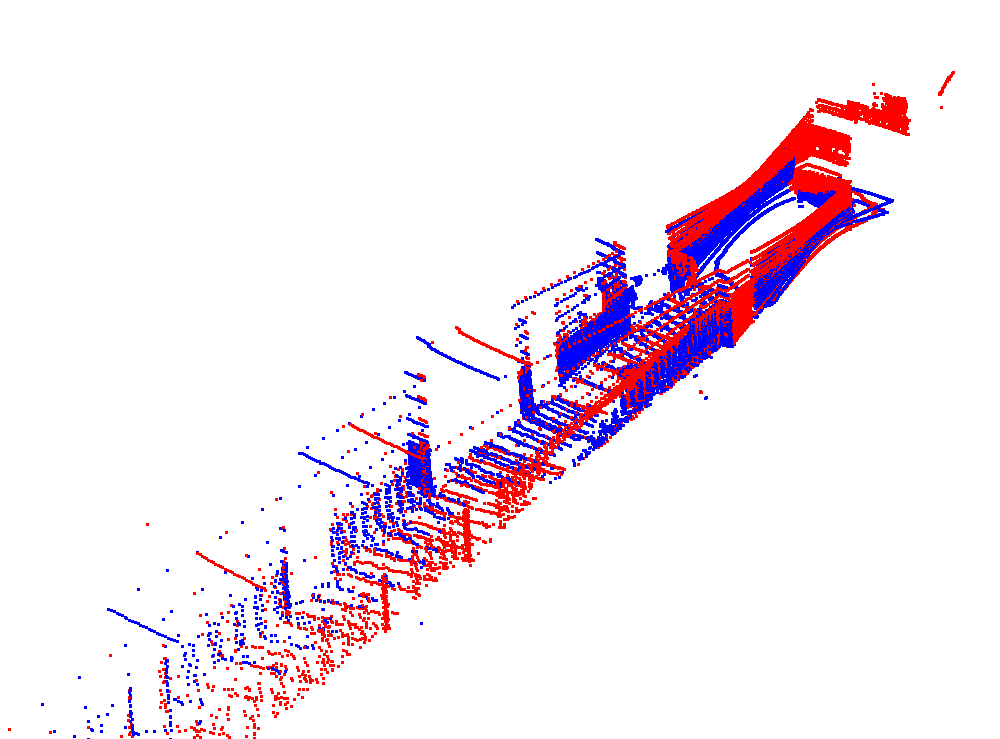}}
		\qquad
		\subfloat[]{%
			\includegraphics[width=0.225\textwidth]{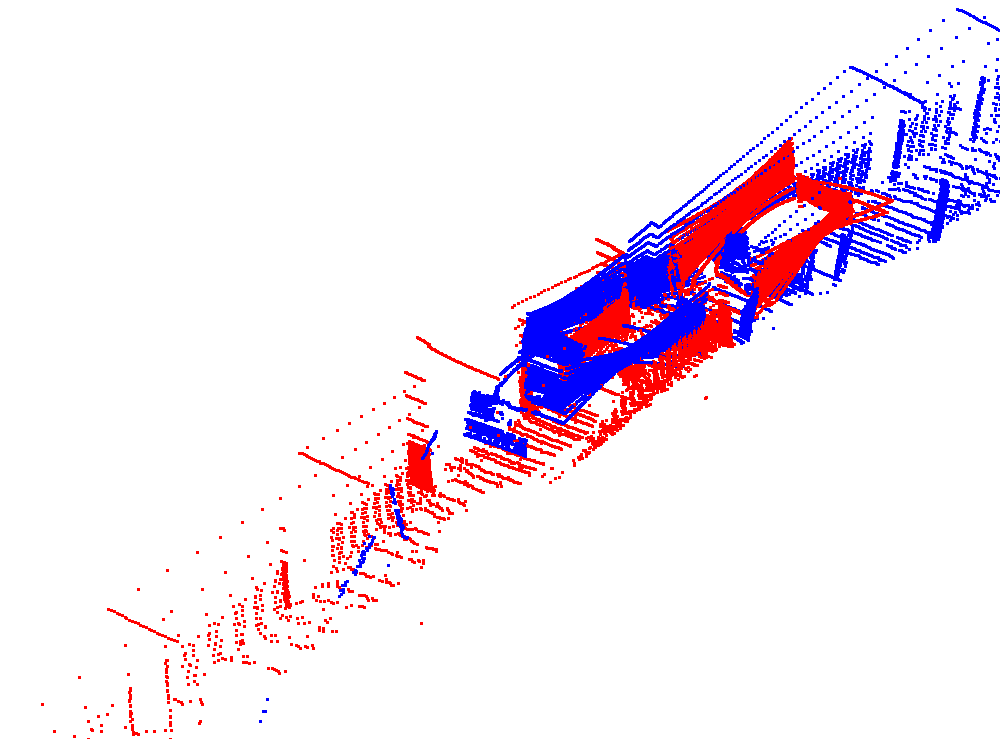}}
		\qquad
		\subfloat[]{%
			\includegraphics[width=0.225\textwidth]{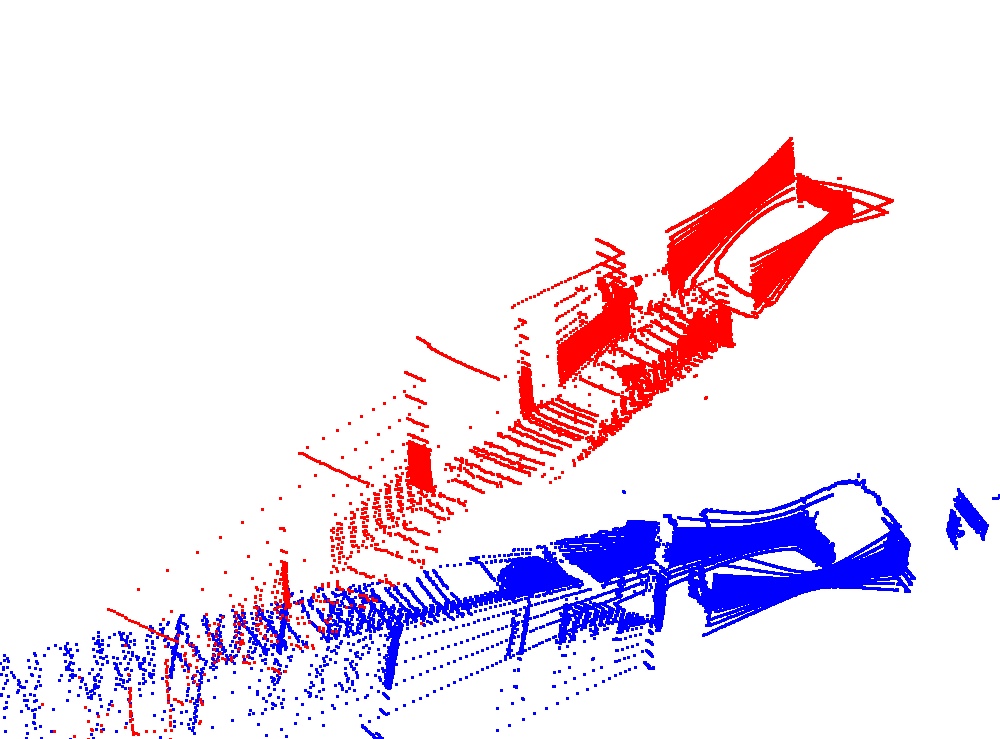}}
		\qquad
		\subfloat[]{%
			\includegraphics[width=0.225\textwidth]{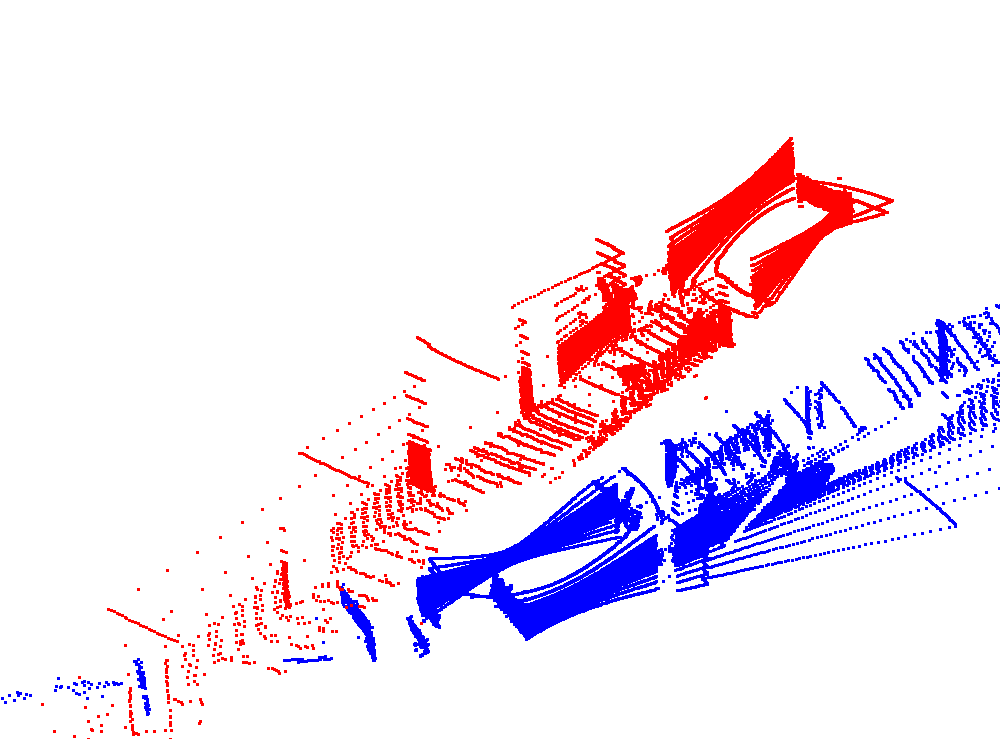}}
		
		\centering
		\caption{Four different correspondent relationships. The red point cloud is the referenced from LiDAR 1, and the blue one is aligned point cloud from \mbox{LiDAR 2}. Fig. \ref{fig.result_a} is a reasonable result and all the others are incorrect results.}
		\label{four_result}
	\end{figure}
\begin{figure}[t]
	\subfloat[]{%
		\label{fig_comp_a}
		\includegraphics[width=0.21\textwidth]{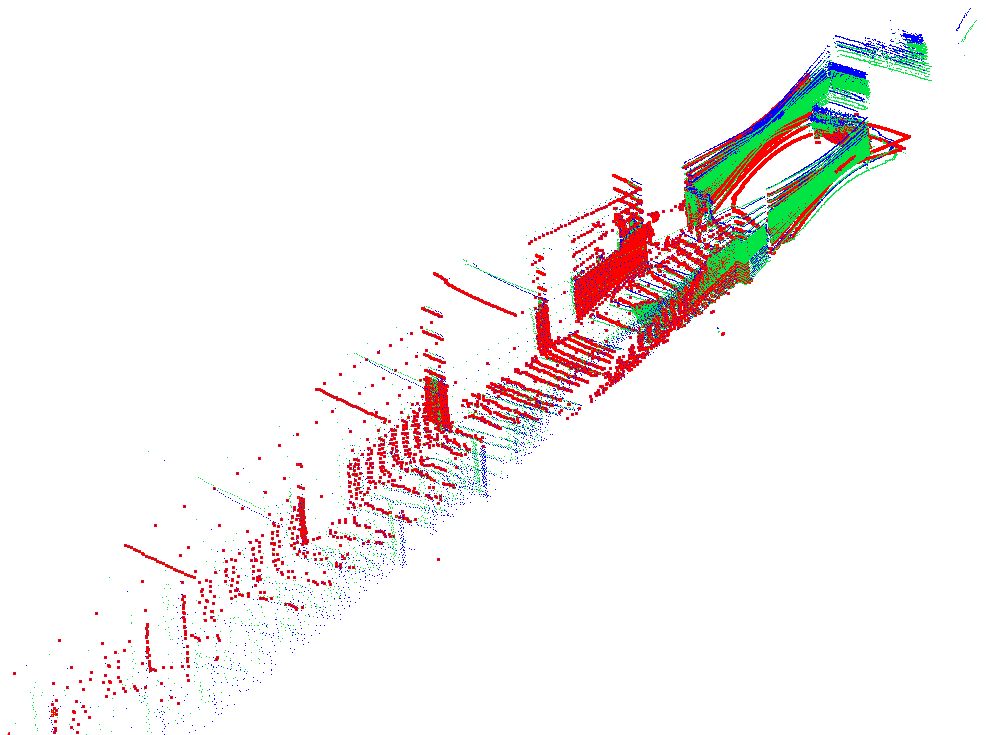}}
	\qquad
	\subfloat[]{%
		\label{fig_comp_b}
		\includegraphics[width=0.21\textwidth]{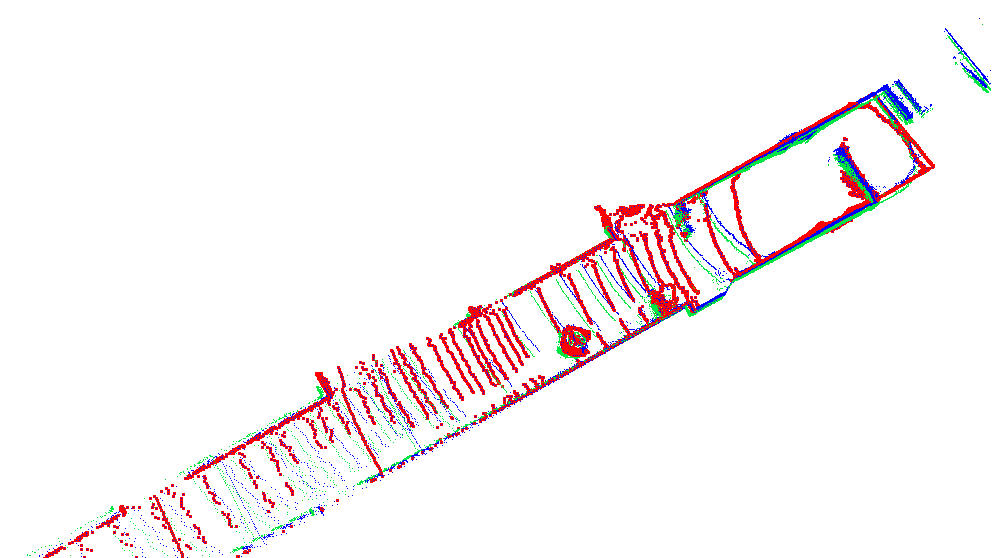}}
	\centering
	\caption{Stacking of different results. The referenced point cloud is red, the green point cloud is generated by planar surfaces approach, and the blue is our method's result. Fig. \ref{fig_comp_a} is the normal view result, Fig. \ref{fig_comp_b} is result of top view. We can see that the results are similar.}
	\label{figure_comparing_result}
			\vspace{-1em}
\end{figure}

	We take a method for comparison. The method is developed based on an algorithm using planar surfaces \cite{jiao2019novel}, and we select a calibration environment in the indoors corridor for calibration. This environment has enough planes perpendicular to each other to perform the planar surfaces approach.
	
	We get several results because different matching relationships lead to different results.
	\mbox{Fig. \ref{four_result}} illustrates the four results which corresponded with \mbox{Fig. \ref{four_axis}}. 
	We use the ICP algorithm to select the most appropriate result. Each pair of red and black point clouds in \mbox{Fig. \ref{four_result}} can be registered by using the ICP algorithm, and then we evaluate the difference between the results and the identity matrix. The results are given \mbox{in Tab. \ref{table_four_result}}. Since a larger RQE represents better calibration, we can see that we have similar results with the state-of-the-art approach.

	\begin{table}[t]
		\caption{The difference between ICP result in \mbox{Fig. \ref{four_result}} and identity matrix}
		\label{table_four_result}
		\begin{center}
			\begin{tabular}{c c c}
				\hline \toprule[0.03cm]
				Number & Rotation Error $e_{r}$ & Translation Error $e_{t}$\\
				\hline \toprule[0.03cm]
				a & 0.1027 & 0.3949\\ 
				b &  0.8393 & 0.9455\\
				c & 1.6875 & 17.4749\\
				d & 0.4021 & 8.4218\\
				\hline \toprule[0.03cm]
			\end{tabular}
		\end{center}
			\vspace{-1.5em}
	\end{table}


	\begin{table}[t]
		\caption{Evaluation of different approaches by RQE}
		\label{table_for_measure}
		\begin{center}				
				\begin{tabular}{c c}
					\hline \toprule[0.03cm]
					Approaches & $\textrm{RQE}$ \\
					\hline \toprule[0.03cm]
					Planar Surfaces &  0.0824 \\
					Proposed & 0.0822 \\
					ICP & 0.0461\\
					Wrong Result &  0.0497\\
					\hline \toprule[0.03cm]
				\end{tabular}
		\end{center}
			\vspace{-1.5em}
\end{table}
	


	\subsection{Discussion}
	We have an assumption in this method: LiDARs share overlapping fields of view. The proposed method may fail in several cases. For instance, if the poles are wrongly detected, the entire system will get the wrong result. Some bands of the LiDAR may get very inaccurate data, and this will impact the results. However, compared with the planar surface approach, there are some advantages of our proposed method:
	\begin{itemize}
		\item Poles can be fitted with very few points, but the state-of-the-art approach requires more points on the surfaces.
		\item Fewer constraints are required, as long as the poles can be detected by two LiDARs.
		\item There is no requirement for how a LiDAR must be installed, and it can be flexibly adapted to various situations. Unlike the planar surfaces approach, it needs to be installed horizontally on a platform.
	\end{itemize}

	
	\section{CONCLUSIONS} \label{seg5}
	In this paper, we presented an automatic approach for calibrating LiDARs with two poles stickered with retro-reflective tape. There are still some problems with this method; for instance, the method of pole recognition depends on the intensity information from the LiDAR and this data is not stable, decreasing as the distance increases. Although in some cases its calibration accuracy is lower than that of other algorithms, it does not depend on the terrain and does not require the platform to move, so calibration tasks can be performed in any scenario, like on an open harbor.
	
	
	\section*{Acknowledgment}
	This work was supported by the National Natural Science Foundation of
	China, under grant No. U1713211, the Research Grant Council of Hong Kong
	SAR Government, China, under Project No. 11210017, No. 21202816, and
	the Shenzhen Science, Technology and Innovation Commission (SZSTI) under
	grant JCYJ20160428154842603, awarded to Prof. Ming Liu.

	\bibliographystyle{ieeetr}

\end{document}